\documentclass[
]{ceurart}

\usepackage{listings}
\usepackage{placeins}
\usepackage{soul, color,wasysym}
\usepackage{balance}
\balance

\begin{document}

\copyrightyear{2024}
\copyrightclause{Copyright for this paper by its authors.
  Use permitted under Creative Commons License Attribution 4.0
  International (CC BY 4.0).}

\conference{CLEF 2024: Conference and Labs of the Evaluation Forum, September 09–12, 2024, Grenoble, France}

%\title{Pushing the Boundaries of AI-Powered Medical Image Synthesis and Prompt Generation: Insights From the MedVQA-GI Challenge Using CLIP, Fine-Tuned Stable Diffusion, and Dream-Booth + LoRA Techniques}

\title{Advancing AI-Powered Medical Image Synthesis: Insights from MedVQA-GI Challenge Using CLIP, Fine-Tuned Stable Diffusion, and Dream-Booth + LoRA}

\title[mode=sub]{Notebook for the CS\_Morgan Lab at CLEF 2024}

\author[1]{Ojonugwa Oluwafemi Ejiga\space    Peter}[%
orcid=0009-0003-2039-3075,
email=ojeji1@morgan.edu,
url=https://ejigsonpeter.github.io/,
]

\cormark[1]
\fnmark[1]
\address[1]{Department of Computer Science, SCMNS School, Morgan State University, Baltimore, Maryland 21251, USA}

\author[1]{Md  Mahmudur Rahman}[%
orcid=0000-0003-0405-9088,
email=Md.Rahman@morgan.edu,
url=https://github.com/rahmanmm,
]
\fnmark[1]

\author[2]{Fahmi Khalifa}[%
email=fahmi.khalifa@morgan.edu,
orcid=https://orcid.org/0000-0003-3318-2851
]

\address[2]{Electrical \& Computer Engineering Dept., School of Engineering, Morgan State University, Baltimore, Maryland 21251, USA}
%% For Dr. Fahmi

\cortext[1]{Corresponding author.}
\fntext[1]{These authors contributed equally.}

\begin{abstract}
   The MEDVQA-GI challenge addresses the integration of AI-driven text-to-image generative models in medical diagnostics, aiming to enhance diagnostic capabilities through synthetic image generation. Existing methods primarily focus on static image analysis and lack the dynamic generation of medical imagery from textual descriptions. This study intends to partially close this gap by introducing a novel approach based on fine-tuned generative models to generate dynamic, scalable, and precise images from textual descriptions. Particularly, our system integrates fine-tuned Stable Diffusion and DreamBooth models, as well as Low-Rank Adaptation (LORA), to generate high-fidelity medical images. The problem is around two sub-tasks namely: image synthesis (IS) and optimal prompt production (OPG). The former creates medical images via verbal prompts, whereas the latter provides prompts that produce high-quality images in specified categories. The study emphasizes the limitations of traditional medical image generation methods, such as hand sketching, constrained datasets, static procedures, and generic models. 
    Our evaluation measures showed that Stable Diffusion surpasses CLIP and DreamBooth + LORA in terms of producing high-quality, diversified images. Specifically, Stable Diffusion had the lowest Fréchet Inception Distance (FID) scores (0.099 for single center, 0.064 for multi-center, and 0.067 for combined), indicating higher image quality. Furthermore, it had the highest average Inception Score (2.327 across all datasets), indicating exceptional diversity and quality. This advances the field of AI-powered medical diagnosis. Future research will concentrate on model refining, dataset augmentation, and ethical considerations for efficiently implementing these advances into clinical practice.

\end{abstract}



\maketitle
\section{Introduction}
Synthetic image generation can be defined as the process of creating fake pictures that are convincing enough to be considered as originals \cite{ref1}. This technology has developed much further and is based on various principles such as Generative Adversarial Networks (GANs), which is a generator and discriminator-based approach that generates fake images and checks their authenticity, and Variational Auto Encoders (AEs) including its improved form VQ-VAE which are better in image generation compared to basic GANs since they produce diverse images and are easier to train \cite{ref1}. One of the most pioneering and widely used techniques in the generation of synthetic images is the GANs first proposed by 
Goodfellow et al \cite{ref2}. GANs consist of two neural networks: an image generator like the concept of GAN and a discriminator which assesses the generated images. This process of training puts the generator in a position to continuously generate image outputs that oppose the discriminator which in return improves its ability to differentiate between the real and synthetic cases. In a longer period, this process is becoming an ATV and can produce a real image that is hardly distinguishable from a synthetic one.

Since the inception of GANs, there have been so many enhancements that have been made to the model. Altered versions like DCGANs \cite{ref3}, cGANs \cite{ref4}, and StyleGANs \cite{ref5} all have improvised the quality as well as the number of images obtained from generative models. DCGANs introduced convolutional layers for the stability of training as compared to a basic GAN, while cGANs contain conditional information like class labels for producing images corresponding to a certain category. Karras et al. \cite{ref5} have continued to take Generative Adversarial Network technology further with StyleGANs, availing the possibility of manipulating the style and features of the resulting images through the style transfer method.

In parallel, the idea of deep generative models based on VAEs has given another strong direction in the generation of synthetic images. Originally put forward in 2013 by Kingma and Welling \cite{ref6}, VAEs have the objective of learning a continuous latent variable model for data generative modeling. Unlike GAN which creates images during the training and passing of an adversarial system, VAE codes the data input into an encoded space and then decodes back to an image. To this end, it enables the change between latent representations easily making it possible for the generation of new and coherent images.

VQ-VAE was developed by Oord et al. as an improvement to the basic framework of VAE \cite{ref7}. The proposed work, VQ-VAE, enhances the original VAEs by implementing vector quantization of the latent space for generating images that are of better quality and have more variation. The quantized nature of the latent space in VQ-VAE is beneficial here to train the network and represent complex image structures better.

\section{Impact of Synthetic Image Generation}
Synthesizing an image comes with many advantages, especially achieving high outcomes when there is little actual image data available. In the vast area of computer vision, obtaining and labeling the vast datasets of real images often becomes a time-consuming, costly, and, at times, even impossible affair. Hence, synthetic image generation can eliminate these issues by making available a large and diverse dataset for training deep learning algorithms \cite{ref1}. A notable use of synthetic image generation is in the improvement of other models, particularly those of machine learning. Real datasets can be complemented by synthetic ones, thus expanding their size and the variability of samples. This augmentation assists in reducing the chances of overfitting, that is, models that show good performance against training data, but poor performance against new data, by the exposure of models to different samples during training. This means that we’re able to get better models that generalize well to other data points. Concerning the application of synthetic image generation in the medical diagnosis setting, the possibilities are enormous. Imaging including X-ray, MRI, and CT scans are critical in diagnosing several illnesses. One of the major issues for obtaining a vast and heterogeneous medical image dataset is the privacy of patients, the cost of the imaging equipment, the time involved, and the expertise needed to annotate the images. 

These issues can be addressed by employing synthetic images as medical images to train diagnosis models based on many photos that do not infringe on the patient's privacy rights. Synthetic picture synthesis can also aid in the creation of talking chatbots. Chatbots can improve the user experience by providing contextual visuals based on user input. For example, a medical chatbot could provide appropriate medical visuals to help patients comprehend and be satisfied with their ailments or operations. Furthermore, synthetic images can be used in search engines to generate copyright-free photos on the fly. When search results for specific photos are restricted, text-to-image generation can help by creating suitable images based on user queries. This can enhance the user experience and provide valuable visual content without infringing on copyright. 

\section{Tasks Performed and Overall Objectives}
The ImageCLEFmed \cite{ref8} MEDVQA-GI challenge is divided into two primary tasks \cite{ref8}: Image synthesis and Optimal Prompt generation.  The dataset used to build the development dataset was gotten from \cite{ref8} \cite{ref9} \cite{ref10}\cite{ref11}. The primary objectives of the experiments performed for these tasks are:
\begin{enumerate}
    \item To generate high-quality, diagnostically relevant medical images from textual descriptions.
    \item To optimize the prompts used for generating these images to ensure they fall within specified categories and maintain high quality and diversity.
    \item To evaluate the effectiveness of these synthetic images in training machine learning models for medical diagnostics
\end{enumerate}

\subsection{Image Synthesis}
\begin{figure}[ht]
    \centering
    \includegraphics[width=0.75\linewidth]{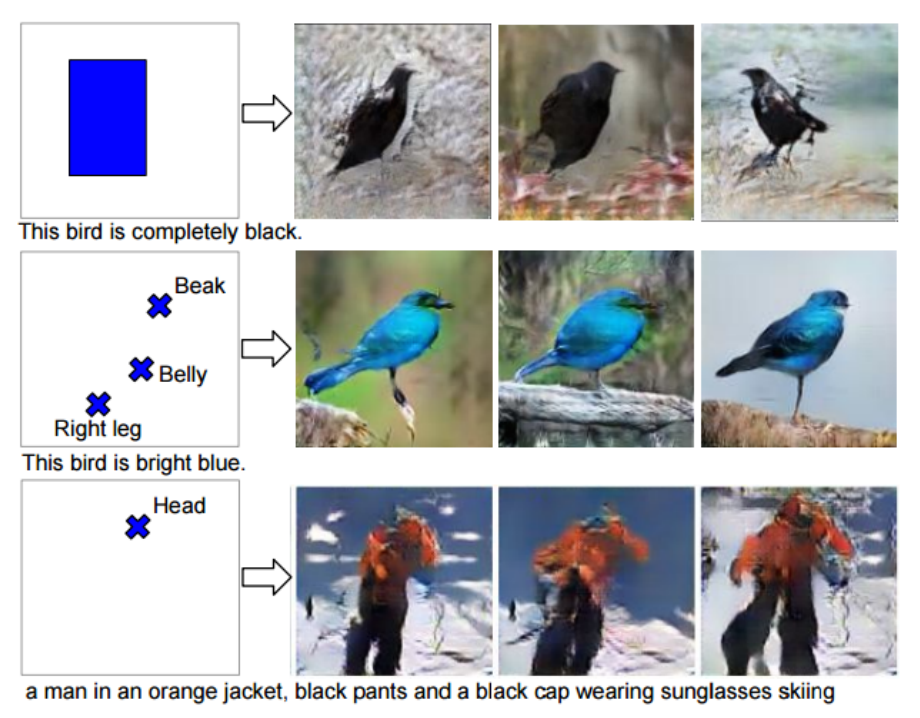}
    \caption{Examples of the  Generative Adversarial What-Where Network (GAWWN) \cite{ref13}.}
    \label{fig:Figure_2}
\end{figure}

Image synthesis is the process of creating an image from an input text, sketch, or other source, such as another image or mask \cite{ref12}. It is an essential problem in the field of computer vision, and the research community has been drawn to attempt to tackle it at a high level to make photorealistic images. In this task, participants have the role of designing images based on the stimuli provided under various categories. This task requires participants to use text-to-image generative models to build a large dataset of medical images based on textual prompts. Based on this description, participants may create an image of an "early-stage colorectal polyp". Participants are given a development dataset comprising prompt-image pairings to help them build their answers, and they are then given prompts to make corresponding images. Figure~\ref{fig:Figure_2} contains some examples of image synthesis with the Generative Adversarial What-Where Network (GAWWN). In GAWWN, images are conditioned on both text descriptions and image positions supplied as key points.

\FloatBarrier

\subsection{Optimized Prompt Generation}
Prompt Engineering is the activity of using text inputs to direct generative AI models, such as text-to-text and text-to-image models, to produce certain results \cite{ref14}. Yaru et al. \cite{ref15} created a prompt adaptation which is a framework that automatically converts user input into model-preferred prompts, reducing human engineering costs and enhancing visuals in text-to-image models. Participants are charged with creating images from their prompts under various categories. The purpose is to create synthetic images capable of training predictive machine learning models. This entails developing prompts that produce visuals of polyps, specific anatomical landmarks, and medical tools, among other things.

\begin{figure}[htbp]
    \centering
    \includegraphics{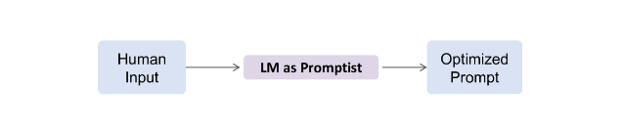}
    \caption{Prompt optimization flow showing the iterative process of refining prompts for improved AI model responses~\cite{ref15}.}
    \label{fig:Figure_3}
\end{figure}
\FloatBarrier

% \section{Main Objectives of Experiments}
% The primary objectives of these experiments are:
% \begin{enumerate}
%     \item To generate high-quality, diagnostically relevant medical images from textual descriptions.
%     \item To optimize the prompts used for generating these images to ensure they fall within specified categories and maintain high quality and diversity.
%     \item To evaluate the effectiveness of these synthetic images in training machine learning models for medical diagnostics
% \end{enumerate}

%\section{Approaches Used and Progress Beyond State-of-the-Art}
\section{Proposed Methodology}
The incorporation of complex generative models into medical imaging has made considerable strides in improving the accuracy and reliability of medical diagnosis. Our approach entails fine-tuning the Stable Diffusion and Dream booth models using Low-Rank Adaptation (LORA), a technique that improves their ability to generate high-fidelity medical images from textual descriptions. We also use the Contrastive Language-picture Pre-training (CLIP) model to improve the contextual understanding and accuracy of picture production. This study discusses the approaches and advances that go beyond the existing state-of-the-art in medical picture production.

\subsection{Stable Diffusion Model and Dream booth Models} 
The generative models of Stable Diffusion and DreamBooth are among the best, especially for generating a diverse and high-quality image. Stable Diffusion stable is the process of diffusing data and stabilizing an image through a series of denoising. This technique is highly effective in medical imaging where it is very important to have well-defined and accurate images. The core concept behind diffusion models is quite straightforward. They use several $T$ steps to gradually add Gaussian noise to the input image ($x0$) \cite{ref16}. Hence, this is called the forward process. Notably, this is not connected to a neural network's forward pass. This part is required to build targets for our neural network (the image after applying $t < T$ noise steps). Then, a neural network is trained to recover the original data by reversing the noise process. By modeling the opposite process, we may generate new data. This is known as the reverse diffusion process, or, more generally, the sampling process of a generative model \cite{ref16}. Again, in contrast to traditional GANs, in Stable Diffusion, the images are fine-tuned probabilistically, not through adversarial learning \cite{ref17}.  

\begin{figure}[ht]
    \centering
    \includegraphics[width=1\linewidth]{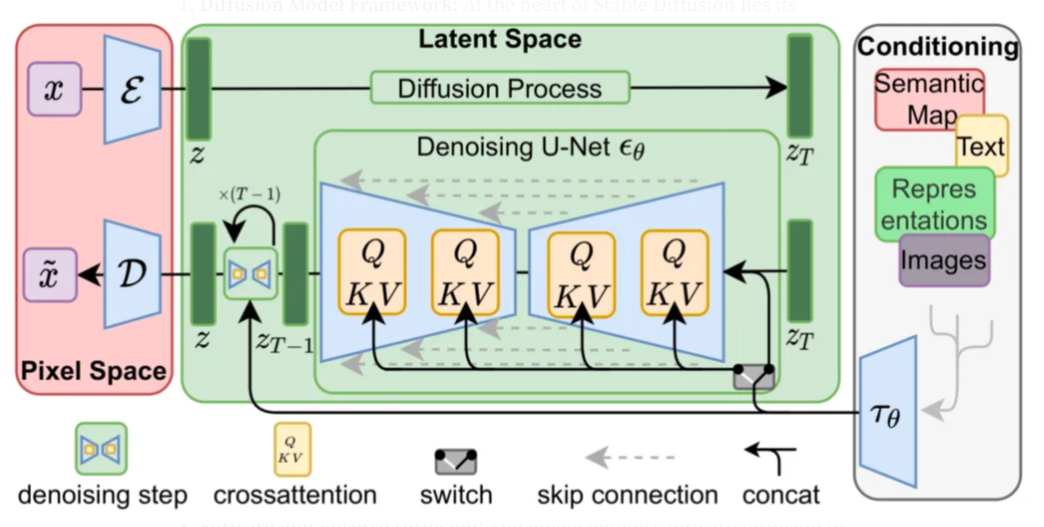}
    \caption{Understanding diffusion process \cite{ref16}.}
    \label{fig:Figure_4}
\end{figure}
\FloatBarrier

A stable diffusion model’s architecture consists of a U-Net, a symmetric architecture with input and output of the same spatial size. The input image is first down-sampled and then up-sampled until reaching its initial size. U-Net consists of Wide ResNet blocks, group normalization, and self-attention blocks. The diffusion timestep $t$ is specified by adding a sinusoidal position embedding into each residual block \cite{ref18}. The diffusion process is divided into two, namely forward and backward diffusion processes \cite{ref19}. The equation below shows the forward process:
\begin{equation}
    dx = f(x,t)dt = g(t)dw
\end{equation}

Below is a backward diffusion process that reverses the time 
\begin{equation}
    dx = f(x,t) - g(t)^2\Delta x logp(x,t)dt + g(t)dw
\end{equation}

Hence, the time-dependent score function $\Delta x logp(x,t)$ is known. Then, the diffusion process can be reversed.  Training a diffusion model entails learning to denoise, for example, if a model can be scored using $f_{\theta}(x,t)\approx\Delta logp(x,t)$ then denoising can be achieved by reversing the diffusion equation $x_2\to x_{t-}$\cite{ref19}. The Score model $f_{\theta}: X \times [0,1]\to X$ can be referred to as a time-dependent vector field over $x$ space. Hence, the Training objective is to first infer noise from a noised sample \cite{ref19} as seen in the equation below:

\begin{equation}
    x~p(x,y),\in ~ N(0,1) t\in[0,1]
    Min||\in f_{\theta}(x+\sigma^2\in,y,t)||_2^2
\end{equation}

Furthermore, adding Gaussian noise $\in$ to an image $x$ with scale $\sigma t$, helps the diffusion model to learn how to infer the noise $\sigma$. 
Another method of inferring noise is through what is known as conditional denoising. This method infers noise from a noised sample, based on a condition y:

\begin{equation}
    x,y~p(x,y),\in~N(0,1),t\in [0,1]
    Min||\in + f_{\theta}(x+\sigma^t\in,y,t)||_2^2
\end{equation}

Hence, the conditional score model $f_\theta: X \times Y \times 0,1,\to X$ uses Unet to model image-to-image mapping while modulating the Unet with condition in the form of a text prompt. For this research, a Stable diffusion model was fine-tuned using the dataset provided by image CLEF \cite{ref8}.  Hence, Stable diffusion is a state-of-the-art generative artificial intelligence model that transforms text-to-image generation by using textual and visual prompts to create one-of-a-kind, lifelike images, films, and animations \cite{ref20}. It was introduced in 2022 and makes use of latent space and diffusion technology to provide an effective and easily accessible means of expressing creativity \cite{ref20}. As a versatile AI model, Stable Diffusion lowers processing needs, enabling users to run on consumer-grade devices with GPUs, and supports multimedia generation beyond static images. As little as five photos are needed for customized results because of its fine-tuning capability. The open license that permits users to use, alter, and redistribute the software freely makes Stable Diffusion accessible and user-friendly.

The stable diffusion model comprises of the following \cite{ref20}:
\begin{enumerate}
    \item \textbf{Diffusion Model Framework}: This framework uses a noise predictor and reverse diffusion process to reproduce the original image. It differs from traditional image creation models by encoding images using Gaussian noise. 
    \item \textbf{Latent Space Magic}:  It Maintains image quality while lowering processing requirements by operating in a latent space with diminished definition. 
    \item \textbf{Architecture}:  The architecture comprises of four (4) key components:
    \begin{enumerate}
        \item \textit{VAE}: This algorithm compresses an image of $512\times 512$ pixels into a $64\times 64$ latent space. 
        \item \textit{Forward and Reverse Diffusion}: Gaussian noise is progressively added in both forward and reverse diffusion until only random noise is present. It contributes to the uniqueness of the images.
        \item \textit{Noise Predictor (U-Net)}: Estimates and subtracts noise from the latent space to improve the visual output. 
        \item \textit{Text Conditioning}: Stable Diffusion uses text prompts to introduce conditioning. Text prompts are analyzed by a CLIP tokenizer, which embeds them into a 768-value vector and uses a text transformer to direct the U-Net noise predictor.
    \end{enumerate}
\end{enumerate}

\begin{figure}[ht]
    \centering
    \includegraphics[width=1\linewidth]{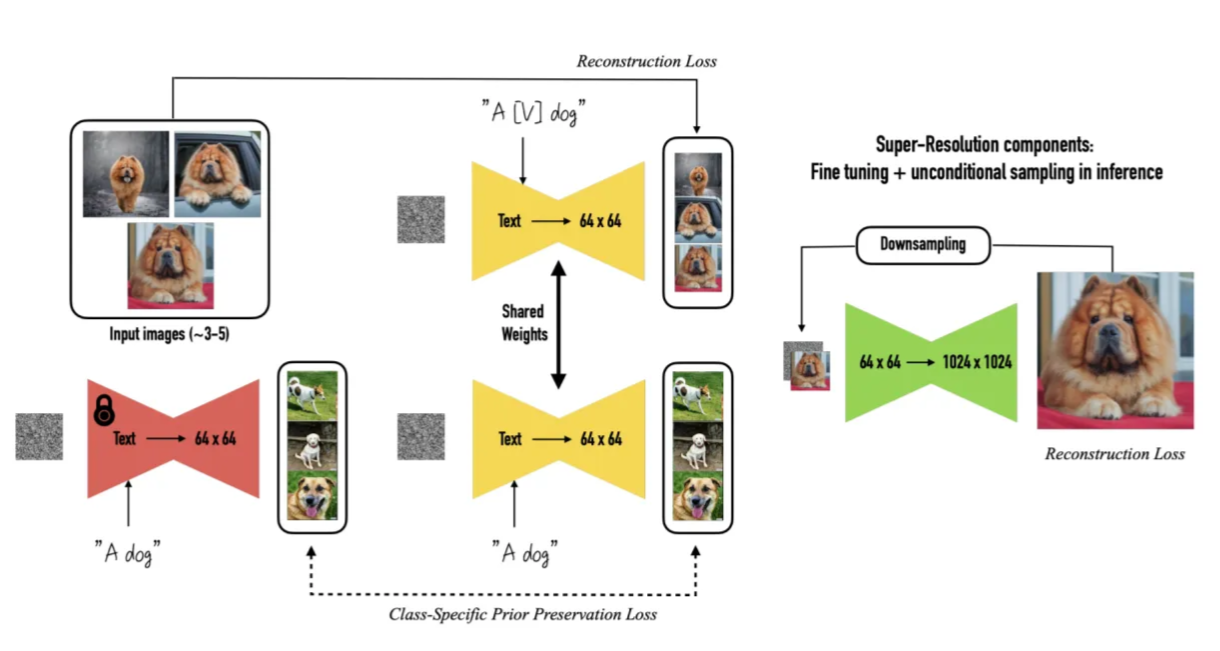}
    \caption{Finetuning stable diffusion \cite{ref20}.}
    \label{fig:Figure_6}
\end{figure}

Fine-tuning a stable diffusion model is crucial in generative AI, as it allows the model to adapt to specific datasets and tasks, improving its effectiveness and aligning with user-defined objectives \cite{ref20}. This process allows the model to capture unique features and patterns, enhancing performance and generating contextually relevant images. It also improves image quality by capturing finer details and allowing for continuous improvement over epochs. Fine-tuning often involves mixed-precision support, ensuring computational efficiency. It also allows customization for specific tasks, ensuring the model remains relevant and adaptable to changing data distributions. Figure \ref{fig:Figure_6} illustrates the process of fine-tuning the stable diffusion model. Hence, the stable diffusion model was fine-tuned using colonoscopy image data and custom prompt \cite{ref8} to generate synthetic colonoscopy images.

\FloatBarrier

DreamBooth on the other hand builds upon these features with domain-specific fine-tuning, which means that the created images are closer to the specified criteria, such as in medical imagery. Thus, with LORA’s help to fine-tune these models for our specific needs, there is enhanced capability to generate medical images that are as close as possible to the textual description, compared to the standard approaches \cite{ref21}. DreamBooth gives users control over the power of stable diffusion, enabling them to fine-tune pre-trained models to produce original images based on their notions. DreamBooth is unique in that it can be customized with a small number of images—usually 10 to 20—making it effective and user-friendly. The main goal of DreamBooth is to impart fresh knowledge to the model, which is accomplished through a procedure known as fine-tuning. \cite{ref22}. This process begins by feeding the concept into an already-existing Stable Diffusion model Figure~\ref{fig:Figure_6} using a set of pictures. This could be anything from pictures of your beloved dog to a particular kind of art. DreamBooth then uses a designated token, usually represented by a 'V' in rectangular braces, to direct the model to produce visuals that correspond with your notion. DreamBooth is very good for subject-driven generation \cite{ref23}.

\subsection{Contrastive Language-Image Pre-training (CLIP)}
CLIP is an extension of a substantial corpus of research on multi-modal learning, natural language supervision, and zero-shot transfer \cite{ref24}. OpenAI's CLIP model makes use of extensive pre-training on a variety of datasets that include both textual descriptions and images. By learning to associate pictures with their matching written descriptions, CLIP offers a strong comprehension of both textual and visual material \cite{ref4}. Our method uses CLIP to fine-tune the textual inputs and guarantee that the images produced are related to the descriptions given and contextually correct. Precision and context are vital in medical imaging; hence this integration is essential.

\begin{figure} [ht]
    \centering
    \includegraphics[width=1\linewidth]{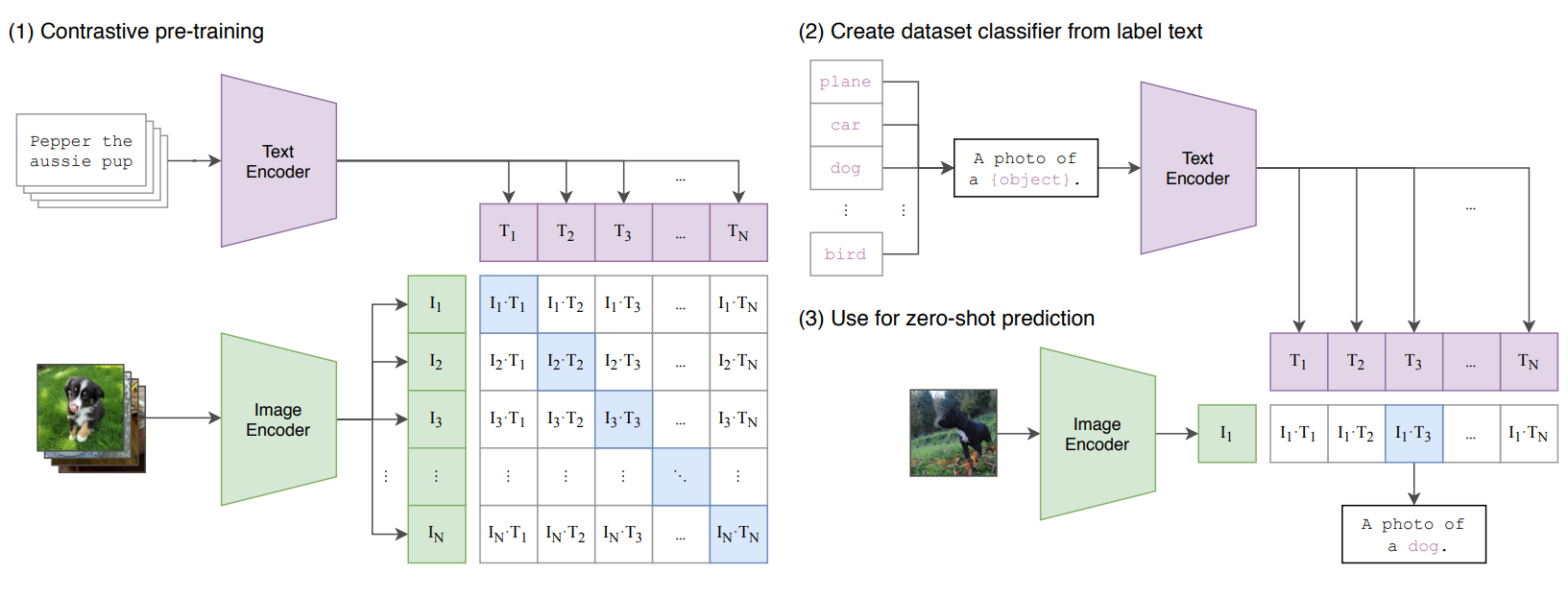}
    \caption{Contrastive Language–Image Pre-training (CLIP) Architecture \cite{ref24}.}
    \label{fig:Figure_7}
\end{figure}
\FloatBarrier

\subsection{Low-Rank Adaptation (LORA)}
LORA is a fine-tuning method that adds a few extra parameters to help huge models be more easily adapted to certain tasks. In the field of medical picture production, where precision and specificity are crucial, this technique is quite beneficial. Low-rank parameter matrices, which LORA introduces, help the pre-trained models adapt to new tasks without requiring a lot of retraining, preserving their efficacy and efficiency \cite{ref25}.  Traditionally, we adjust the weights of a neural network that has already been trained to suit a new task. This modification entails changing the network's initial weight matrix (W). The updated weights can be stated as $(W + \Delta W)$ since the adjustments made to (W) during fine-tuning are collectively represented by $(\Delta W)$\cite{ref26}.  Now, the LoRa technique aims to decompose $(\Delta W)$, instead of directly changing (W). Reducing the computational burden related to fine-tuning large models requires this decomposition. LoRa is a model training technique that solely utilizes lower-rank matrices to enhance the speed and efficiency during model training. Conventional fine-tuning involves retraining the entire model from scratch, an iterative process that may be expensive \cite{ref27}. Compared with LORA, it focuses on modifying a smaller number of parameters to cut computational and memory overhead. By breaking up large weight matrices into matrices of a smaller size, LoRa increases possible values. There is a sharp difference in weights or trainable parameters and that took only 5 million trainable parameters instead of 175 billion. The weights are added in parallel: the new weights are added on top of the pre-existing weights without introducing any extra latency. Since LoRa is effective when used in matrix multiplication, it can help many more use cases, making it a diverse technique \cite{ref27}.

It is also used in image models, such as the Stable Diffusion which utilizes the lower-rank matrices trained on a smaller data sample. These or lower-rank ones can be burned and installed on top of the base Stable Diffusion model to produce style-related stimuli. Some typical applications of LoRa Stable Diffusion models include developing specific styles, creating specific characters, as well as enhancing quality. It appears that the outputs of many LoRAs can lead to specializations, and the various combinations obtained suggest distinctive types of outputs. Hence, LoRa has various benefits for adapting giant models, including performance, accuracy, and adaptability.

\begin{figure}[ht]
    \centering
    \includegraphics[width=0.75\linewidth]{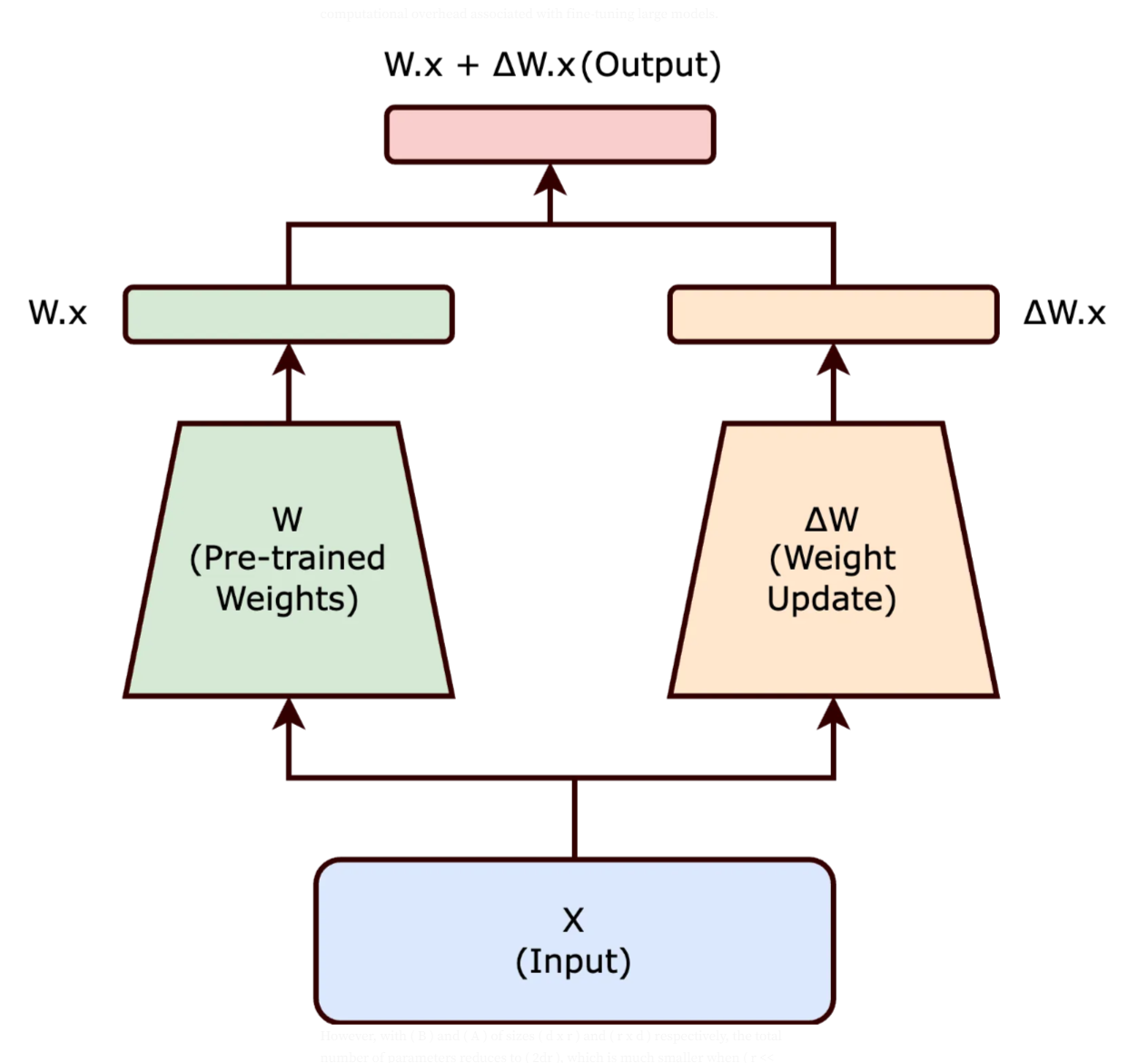}
    \caption{LoRa representation \cite{ref26}.}
    \label{fig:Figure_8}
\end{figure}
\FloatBarrier

\section{Results}
\subsection{Dataset Description and Resources Employed}
This study uses the MEDVQA-GI challenge dataset \cite{ref8}, which is divided into training and test sets. 20242 prompts in a CSV file that is matched to a folder holding 2,000 colonoscopy images that make up the training set. To train the generative models, a rich collection of text-image pairs is made available by the fact that every image in the training set includes several related prompts. 5,000 prompts without accompanying visuals are present in the test set, however. To demonstrate the trained models' ability to synthesize fresh medical images from textual descriptions, the aim is to generate images for each of these 5,000 prompts. 
This research was carried out using the Google Cloud Platform (GCP) Vertex AI Workbench. The specific requirements are given in the Table~\ref{tabSepcs}.

\begin{table}[ht]
\caption{Detailed specific requirements for the resources employed.}
\begin{tabular}{lll}
\toprule
Category & Resource               & Specification                                              
\\ \midrule
\multirow{8}{*}{Hardware} & Machine Type           & g2-standard-48 (Graphics Optimized: 4 NVIDIA L4 GPU, 48 vCPUs, 192GB RAM) \\
        & Data Disk Size         & 100 GB                                                                    \\
        & Boot Disk Size         & 150 GB                                                                    \\
        & External IP            & Enabled                                                                   \\
        & Realtime Collaboration & Disabled                                                                  \\
        & Secure Boot            & Disabled                                                                  \\
        & vTPM                   & Enabled                                                                   \\
        & Integrity Monitoring   & Enabled                                                                   \\ \midrule
\multirow{11}{*}{Software} & Programming Language   & Python                                                  \\
        & Cloud Platform         & Google Cloud (Vertex AI Workbench)                                        \\ \cmidrule{2-3}
        & \multirow{8}{*}{Libraries}              & accelerate\textgreater{}=0.16.0                         \\
        &                       & torchvision                                                               \\
        &                       & transformers\textgreater{}=4.25.1                                         \\
        &                       & datasets\textgreater{}=2.19.1                                             \\
        &                       & ftfy                                                                      \\
        &                       & tensorboard                                                               \\
        &                       & Jinja2                                                                    \\
        &                       & peft==0.7.0                                                               \\ \bottomrule
\end{tabular}
\label{tabSepcs}
\end{table}
\FloatBarrier

\begin{figure}[ht]
    \centering
    \includegraphics[width=1\linewidth]{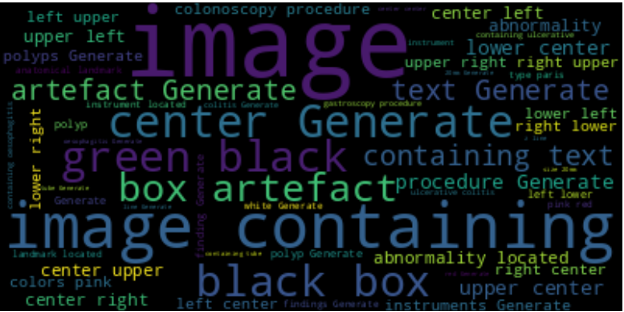}
    \caption{Word Cloud depicting frequently occurring terms in medical imaging analysis, with emphasis on colonoscopy procedures. Prominent words include "image," "colonoscopy," "polyp," and "abnormality," reflecting the focus on diagnostic imaging in gastrointestinal examinations.}
    \label{fig:Figure_9}
\end{figure}

Figure~\ref{fig:Figure_9} is a word cloud that seems to be derived from a corpus of text data about the analysis of medical imagery, with a particular emphasis on colonoscopy procedures. The size of the words within the cloud is indicative of their relative frequency of occurrence within the source text. Featured terms that stand out are "image," "containing," "black box artifact," "generate," "text," "polyp," "abnormality," "instrument," and "colonoscopy procedure." 

\FloatBarrier

\begin{figure}[ht]
    \centering
    \includegraphics[width=1\linewidth]{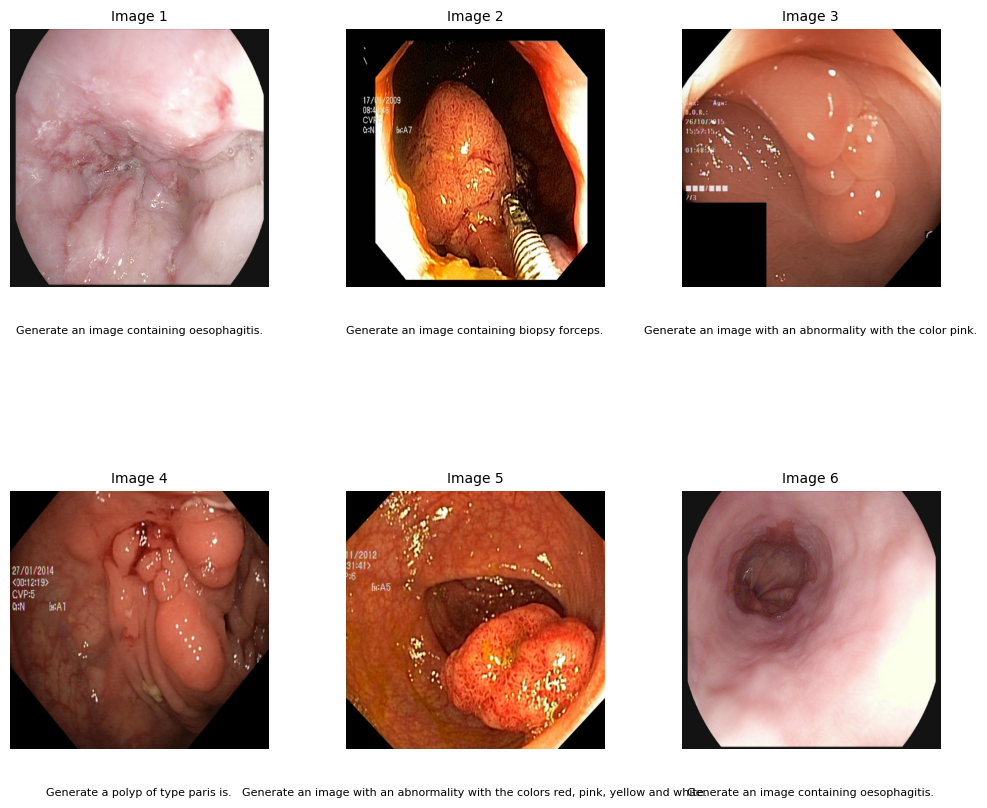}
    \caption{Dataset samples from Hicks et.al \cite{ref8}.}
    \label{fig:Figure_10}
\end{figure}

These terms imply that the underlying text data includes annotations of pathological conditions and medical equipment seen during colonoscopy examinations, instructions for creating synthetic images, spatial localization of findings, and descriptions of image contents. Overall, the word cloud provides a concise representation of the main concepts and vocabulary present in the text corpus, which appears to be devoted to the analysis and interpretation of colonoscopy images, with a particular focus on the identification of lesions, artifacts, and other relevant features. A selection of 12 original colonoscopy pictures from a dataset is shown in Figure \ref{fig:Figure_10}. These pictures show a variety of colon problems and findings, including ulcerations, polyps, inflammation, and lesions. Analysis and research of various colonoscopy instances and observations are made possible by the captions that appear beneath each image. These captions offer instructions or descriptions about the visual material.
\FloatBarrier

\subsection{Contrastive Language-Image Pre-training}
\begin{figure}[!htbp]
    \centering
    \begin{minipage}[b]{0.47\linewidth}
        \centering
        \includegraphics[width=\linewidth]{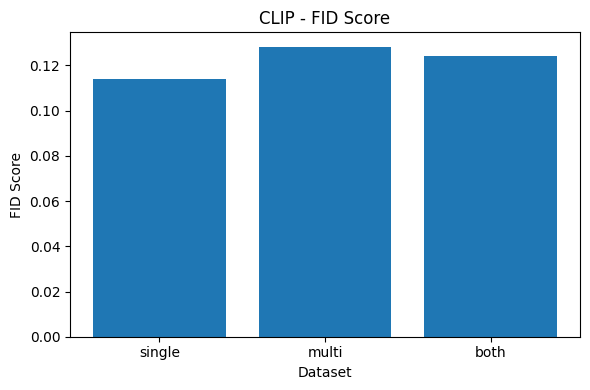}
        \caption{CLIP FID results}
        \label{fig:Figure_11}
    \end{minipage}
    \hfill
    \begin{minipage}[b]{0.47\linewidth}
        \centering
        \includegraphics[width=\linewidth]{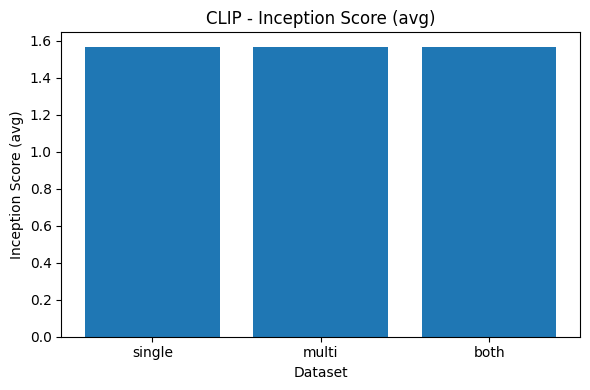}
        \caption{CLIP Inception Scores}
        \label{fig:Figure_12}
    \end{minipage}
\end{figure}

CLIP has the highest Fréchet Inception Distance (FID) scores among the three models, with 0.113972963 for the single-center dataset, 0.128217653 for the multi-center dataset, and 0.124014572 for both datasets combined. These higher scores indicate that the images generated by CLIP have a lower level of similarity to the real images compared to the other models, indicating lower image quality and realism. Table~\ref{tabCLIP} shows the quantitative results for CLIP, which is also visualized in Figure~\ref{fig:Figure_11} and Figure~\ref{fig:Figure_12}. The higher FID score for the multi-center dataset implies that CLIP struggles to generate realistic images when trained on a diverse dataset from multiple medical centers. The overall high FID scores indicate that CLIP is not the most suitable model for generating high-quality medical images in this context.

\begin{table}[ht]
\caption{Quantitative CLIP results.}
\label{tabCLIP}
\begin{tabular}{lllll}
\hline
Task    & Dataset & Submission & FID Score & Inception Score (avg) \\
\hline
task\_1 & single  & CLIP       & 0.113973  & 1.567673087           \\
task\_1 & multi   & CLIP       & 0.128218  & 1.567673087           \\
task\_1 & both    & CLIP       & 0.124015  & 1.567673087 \\
\hline
\end{tabular}

\end{table}
\FloatBarrier

\FloatBarrier

% \begin{figure}
%     \centering
%     \includegraphics{images/Figure_11.png} 
%     \caption{Fréchet Inception Distance (FID)}
%     \label{fig:Figure_11}
% \end{figure}
% \FloatBarrier
% \begin{figure}
%     \centering
%     \includegraphics{images/Figure_12.png}\\
%     \caption{Inception Score}
%     \label{fig:Figure_12}
% \end{figure}
% \FloatBarrier
\begin{figure}[ht]
    \centering
    \includegraphics[width=0.75\linewidth]{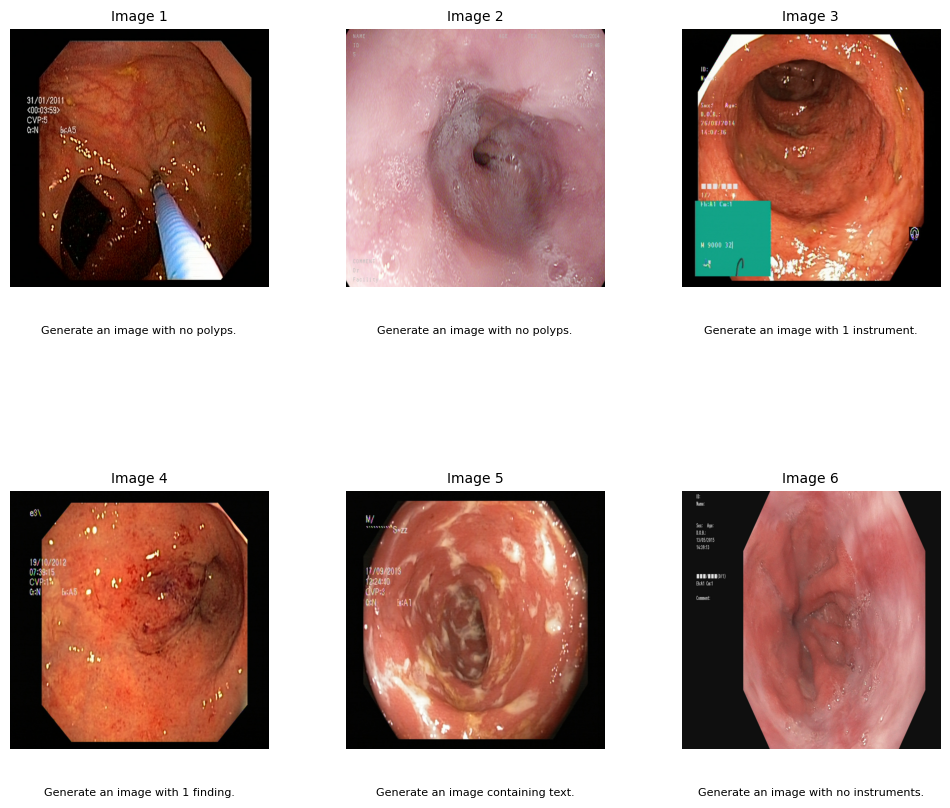}
    \caption{Sampled Synthetic Images from CLIP.}
    \label{ffig:Figure_13}
\end{figure}
\FloatBarrier

CLIP has the lowest average Inception Score among the three models, with a consistent score of 1.567673087 across all datasets (single-center, multi-center, and both). These lower scores indicate that the images generated by CLIP have a lower level of diversity and quality compared to the other models. The consistency of the scores across different datasets indicates that CLIP's performance is not significantly affected by the dataset's origin. However, the low Inception Scores raise concerns about CLIP's ability to generate diverse and high-quality medical images in this context. The results indicate that CLIP is not the most suitable model for generating visually coherent and diverse medical images, and further improvements or alternative approaches may be necessary.

\subsection{Fine-tuned Stable Diffusion Results}

\begin{table}[ht]
\caption{Quantitative stable diffusion model results.}
\label{tabStabDiff}
\begin{tabular}{lllll}
\hline
Task    & Dataset & Submission                  & FIDScore & Inception Score (avg) \\
\hline
task\_1 & single  & Finetunned Stable Diffusion & 0.099407                               & 2.326531              \\
task\_1 & multi   & Finetunned Stable Diffusion & 0.064354                               & 2.326531              \\
task\_1 & both    & Finetunned Stable Diffusion & 0.066756                               & 2.326531   \\
\hline
\end{tabular}
\end{table}
\FloatBarrier

\begin{figure}[ht]
    \centering
    \begin{minipage}[b]{0.48\textwidth}
        \centering
        \includegraphics[width=\textwidth]{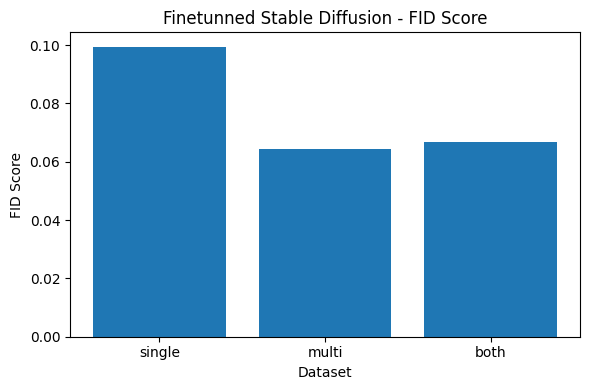}
        \caption{Stable Diffusion FID  results}
        \label{fig:Figure_14}
    \end{minipage}
    \hfill
    \begin{minipage}[b]{0.48\textwidth}
        \centering
        \includegraphics[width=\textwidth]{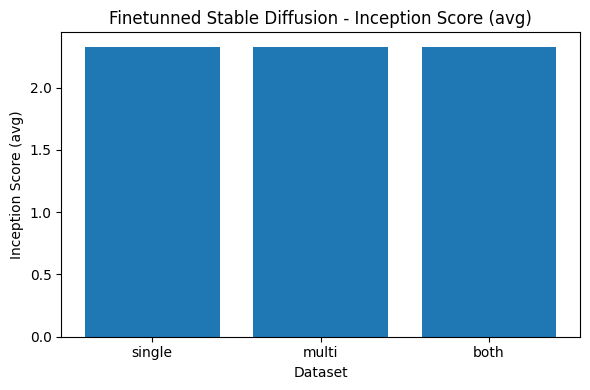}
        \caption{Stable Diffusion Inception Scores }
        \label{fig:Figure_15}
    \end{minipage}
   
\end{figure}
\FloatBarrier
% \begin{figure}[ht]
%     \centering
%     \includegraphics{images/Figure_14.png}
%     \caption{Stable Diffusion FID}
%     \label{fig:Figure_14}
% \end{figure}
% \FloatBarrier

% \begin{figure}[ht]
%     \centering
%     \includegraphics{images/Figure_15.png}
%     \caption{Stable Diffusion Inception Score}
%     \label{fig:Figure_15}
% \end{figure}
% \FloatBarrier

Among the three models, Fine-tuned Stable Diffusion obtains the lowest FID scores: 0.099406576 for the single-center dataset, 0.064354327 for the multi-center dataset, and 0.066755556 for the combined datasets. These low FID ratings imply that, in comparison to the other models, Stable Diffusion produces images that are more realistic and of higher quality. The multi-center dataset's lower FID score suggests that Stable Diffusion can provide more realistic images across many medical centers since it benefits from training on a broad dataset. All things considered; the findings show that the best model for producing high-quality medical images is Fine-tuned Stable Diffusion. Table 3 shows the result for stable diffusion.

Fine-tuned Stable Diffusion achieves an average Inception Score of 2.326530933 across all datasets (single-center, multi-center, and both). While slightly lower than Fine-tuned DreamBooth + LoRa, these scores still indicate that the generated images have a good level of diversity and quality. The consistency of the scores across different datasets suggests that Stable Diffusion can maintain image diversity and quality regardless of the dataset's origin. However, the slightly lower scores compared to DreamBooth + LoRa may indicate that there is some room for improvement in terms of image diversity and quality. Overall, the results demonstrate that Fine-tuned Stable Diffusion is a strong contender for generating diverse and high-quality medical images.

\begin{figure}[ht]
    \centering
    \includegraphics[width=1\linewidth]{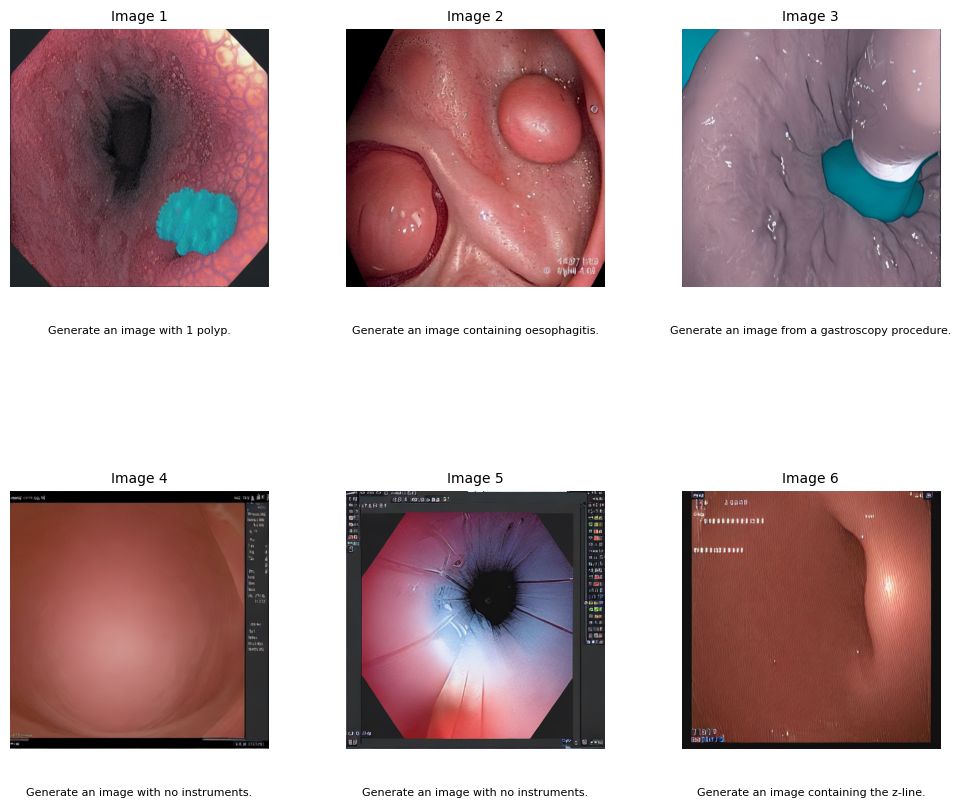}
    \caption{Generated Images from Stable Diffusion.}
    \label{fig:Figure_16}
\end{figure}
\FloatBarrier
\subsection{DreamBooth + LoRA}
The single-center dataset's FID scores for Finetuned DreamBooth + LoRa are 0.110014109, the multi-center dataset's FID scores are 0.072790998, and the combined dataset’s FID scores are 0.075794642. These scores show that although there is still a need for development in terms of image quality and realism, the produced photos do resemble genuine images to some extent. The multi-center dataset's lower FID score indicates that the model performs better when trained on a more varied dataset. However, when compared to fine-tuned Stable Diffusion, the overall FID scores are higher, suggesting that DreamBooth + LoRa might not be as successful in producing high-quality medical images.

\begin{table}[ht]
\caption{Quantitative DreamBototh+LoRA results.}
\label{tabStabDiff1}
\begin{tabular}{lllll}
\hline
Task    & Dataset & Submission                   & FID Score & Inception Score (avg) \\
\hline
task\_1 & single  & Fine-tuned Dreambooth + LoRa & 0.110014                               & 2.36157               \\
task\_1 & multi   & Fine-tuned Dreambooth + LoRa & 0.072791                               & 2.36157               \\
task\_1 & both    & Fine-tuned Dreambooth + LoRa & 0.075795                              & 2.36157   \\
\hline
\end{tabular}
\end{table}
\FloatBarrier

\begin{figure}[ht]
    \centering
    \begin{minipage}[b]{0.48\textwidth}
        \centering
        \includegraphics[width=\textwidth]{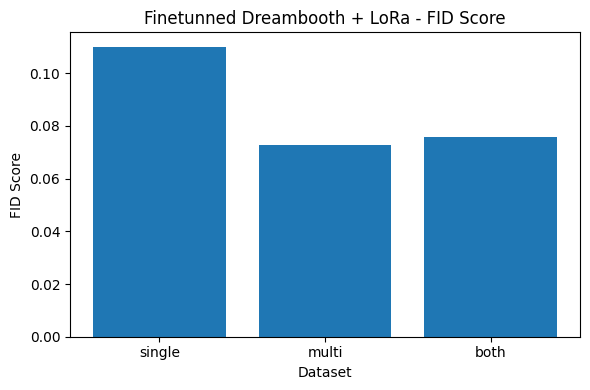}
        \caption{DreamBooth + LoRa  FID  results}
        \label{fig:Figure_17}
    \end{minipage}
    \hfill
    \begin{minipage}[b]{0.48\textwidth}
        \centering
        \includegraphics[width=\textwidth]{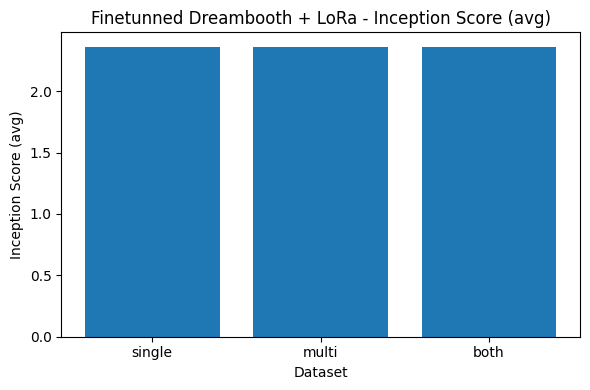}
        \caption{DreamBooth + LoRa  Inception Scores }
        \label{fig:Figure_18}
    \end{minipage}
   
\end{figure}
\FloatBarrier

% \begin{figure} [ht]
%     \centering
%     \includegraphics{images/Figure_17.png}
%     \caption{DreamBooth + LoRa FID}
%     \label{fig:Figure_17}
% \end{figure}
% \FloatBarrier

% \begin{figure} [ht]
%     \centering
%     \includegraphics{images/Figure_18.png}
%     \caption{DreamBooth + LoRa Inception Score}
%     \label{fig:Figure_18}
% \end{figure}
\FloatBarrier
The average Inception Scores for Finetuned Dreambooth + LoRa are consistently 2.361569881 across all datasets (single-center, multi-center, and both). These scores indicate that the generated images have a good level of diversity and quality, as higher Inception Scores are associated with more visually coherent and diverse images. The fact that the scores are consistent amongst datasets implies that DreamBooth + LoRa can preserve image diversity and quality irrespective of the source of the dataset. Though the Inception Scores are high, it's crucial to remember that they are marginally lower than those of Fine-tuned Stable Diffusion, indicating that there could be some space for improvement in terms of picture diversity and quality.

\begin{figure}[ht]
    \centering
    \includegraphics[width=1\linewidth]{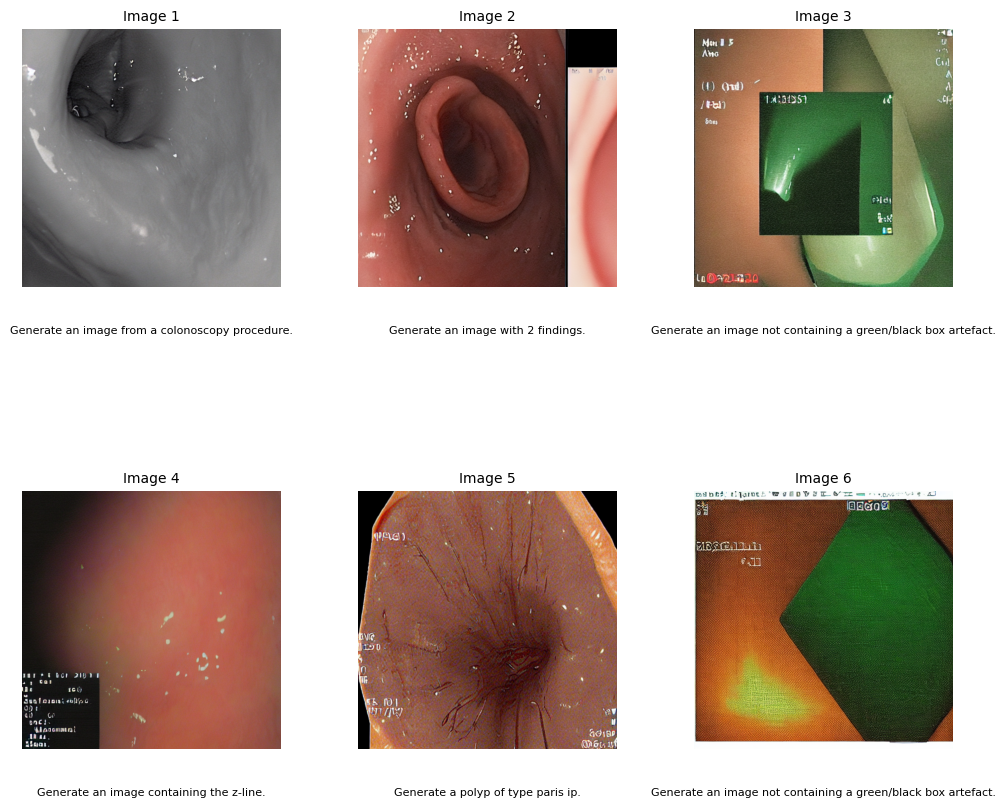}
    \caption{DreamBooth + LoRa Synthetic Images.}
    \label{figDreamBoothLora}
\end{figure}
\FloatBarrier

\section{Result Discussion}
\begin{figure}[ht]
    \centering
    \begin{minipage}[b]{0.48\textwidth}
        \centering
        \includegraphics[width=\textwidth]{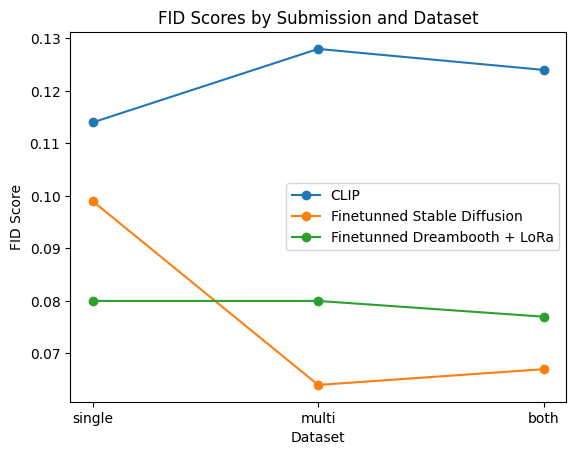}
        \caption{FID scores across three (3) datasets.}
        \label{fig:Figure_20}
    \end{minipage}
    \hfill
    \begin{minipage}[b]{0.48\textwidth}
        \centering
        \includegraphics[width=\textwidth]{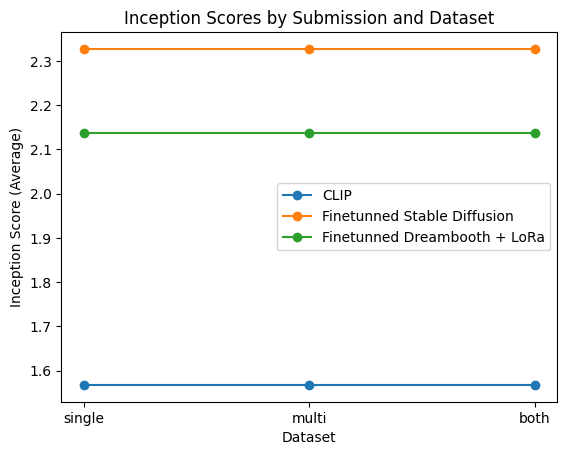}
        \caption{Inception Scores (IS) across three (3) datasets. }
        \label{fig:Figure_21}
    \end{minipage}
   
\end{figure}

The research focuses on the issue of utilizing AI-generated text-to-image generative models in diagnostics. The study evaluates three different approaches: CLIP (Contrastive Language–Image Pre-training), Finetuned Stable Diffusion and Finetuned DreamBooth + LoRa, the metrics used are Fréchet Inception Distance (FID) and Inception Score (IS). As shown by outcomes in Figure \ref{fig:Figure_20}  and Figure \ref{fig:Figure_21}, Finetuned Stable Diffusion yields the smallest FID scores, which means that the generated images are of higher quality and are closer to real-life images than images obtained by other methods. Stable Diffusion also yields the lowest FID scores in all experiments: single center; multi-center; and both datasets with FID scores between 0. 064 to 0. 099. This indicates that Stable Diffusion is superior to CLIP and Finetuned DreamBooth + LoRa and yields more realistic images that resemble real medical images. Consequently, about the Inception Scores, the finetuned DreamBooth + LoRa model attains the optimal average scores, and they always remain at 2.362 across all datasets. This indicates that the produced images come from diverse and reasonably good quality when using DreamBooth + LoRa. However, Finetuned Stable Diffusion closely follows with an average Inception Score of 2.327, suggesting that it also generates diverse and high-quality images. CLIP, on the other hand, has the lowest average Inception Score (1.568), raising concerns about its ability to generate visually coherent and diverse medical images.

\FloatBarrier
% \begin{figure}
%     \centering
%     \includegraphics{images/Figure_20.png}
%     \caption{Submission FID}
%     \label{fig:Figure_20}
% \end{figure}
% \begin{figure}
%     \centering
%     \includegraphics{images/Figure_21.png}
%     \caption{Submission Inception Score}
%     \label{fig:Figure_21}
% \end{figure}

It can also be noticed that the scores received for all the examined approaches are quite similar in single-center, multi-center, and both types of datasets, which indicates that the performance of the developed models does not depend on the dataset’s origin. Nevertheless, it is practical to point out that, in most cases, all three methods can yield higher FID scores in the multi-center dataset, and thus it may be surmised that if the training data is collected from multiple medical centers, it enhances the quality and realism of the generated images. However, the study has some limitations that must be taken into consideration Throughout the research, the real-life applicability of deep learning techniques in the medical field is shown in aspects such as medical image synthesis using artificial intelligence-driven generative models. While IS and FID are the standard tools to assess image quality and realism and can offer some information about the variety of images, they do not give information about how suitable the generated images are for medical applications or can be used by doctors to diagnose patients. Future studies should also evaluate the images generated by the model by experts and introduce the impact of the generated images on the diagnostic indicators.

\section{Challenges and Ethical Considerations}
Conversely, as with most positive impacts, synthetic image generation is also accompanied by several challenges and ethical concerns. Another consideration, which has been raised in other papers, is the potential for resulting images to be fake or misleading. While achieving realistic and synthetic images provides new opportunities for the perfect manipulation of reality, this can be used for the dissemination of fake news, manipulation of the population, or even the creation of fake lab scans. Preventing such misuse is very important, and so it is very important to ensure responsible use of this technology.

Therefore, in the medical domain, the preservation of realistic synthetic images is a matter of concern. The training and diagnostic models can be supplemented with synthetic images, and this is possible only if these images realistically simulate medical conditions. If the synthetic images were not produced properly, they could bring down the quality of the dataset used for the training, which in turn impairs the performance of the model and harms the health of patients by delivering the wrong diagnosis and treatment. Ethical concerns also can be seen in aspects such as the patient’s privacy and protection of their information. The creation of synthetic medical images should be done with proper consideration of patient identifiable features and follow the recommendation of the data protection laws. 

\section{Contributions and Future Work} 
The following are benefits arising from this work since it is a pioneering study that fills a gap in medical image generation and diagnostics. First, using LORA, we want to show that Stable Diffusion and DreamBooth generative models are incredibly effective to use in generating accurate medical images. Future work will discuss the utilization of the generated synthetic images in training diagnostic models while assessing their effect on significant model performance, including generalization performance. Future works the paper will also discuss the ethics involved in the generation of fake images and its possible pitfalls in giving direction to professionals and the public on when and how synthetic images should be used in medicine and other disciplines.

\section{Conclusion}
An important development in artificial intelligence and machine learning is the capacity to produce synthetic visuals from textual descriptions. This technology offers a scalable and privacy-preserving method of producing diverse and high-quality image datasets, which has the potential to revolutionize several industries, most notably medical diagnostics. But technology also presents difficulties and moral dilemmas that need to be handled with care. By utilizing cutting-edge generative models to produce high-fidelity medical images and optimize prompts for high-quality image generation, this project seeks to investigate these capabilities and problems within the framework of the MEDVQA-GI challenge. Our objective is to promote AI-driven medical diagnostics and other applications by resolving the drawbacks of conventional approaches and showcasing the possibilities of synthetic picture production.

\section{Acknowledgement}
This work was supported by the National Science Foundation (NSF) grant (ID. 2131307) “CISE-MSI: DP: IIS: III: Deep Learning-Based Automated Concept and Caption Generation of Medical Images Towards Developing an Effective Decision Support.”

\tolerance=1000
\emergencystretch=\maxdimen
\bibliography{references}

\end{document}